\documentclass{IOS-Book-Article}

\usepackage[T1]{fontenc}
\usepackage{mathptmx}
\usepackage{url}
\usepackage{graphicx}
\usepackage{booktabs}
\usepackage{hyperref}

\def\hb{\hbox to 11.5 cm{}}

\begin{document}

\pagestyle{headings}
\def\thepage{}
\begin{frontmatter}

\title{Towards More Granular Reasoning: Mapping the DSA Transparency Database Taxonomy to Platform Terms and Conditions Using
Large Language Models}

%\markboth{}{December 2026\hb}

\author[A,B]{\fnms{Alyson} \snm{Ranucci}\orcid{0009-0005-2077-8412}},
\author[B]{\fnms{Eline} \snm{Leijten}\orcid{0009-0006-9917-2372}}
and 
\author[A]{\fnms{David} \snm{Graus}\orcid{0000-0002-6245-0870}}

\address[A]{University of Amsterdam, The Netherlands}
\address[B]{Vrije Universiteit Amsterdam, The Netherlands}

\begin{abstract}
Article 24(5) of the Digital Services Act (DSA) requires providers of online platforms to submit every content moderation decision they make, with a corresponding Statement of Reasons (SoR), to the DSA Transparency Database
(DSA-TDB). Such decisions are classified against seventeen high-level attribute categories that categorize the primary reason for the moderation decision. Because over 99.8\% of statements cite a
violation of the provider's own terms and conditions (which is a residual high-level category attribute choice) rather than a legal provision, the database is
overwhelmingly a record of contractual enforcement using catch-all reporting; however, whether its categories actually describe those contractual rules well has not been examined. In this paper, we map the terms and conditions (T\&C) of 31 designated online services, grouped into six service types, against the DSA-TDB's high-level category taxonomy using a few-shot GPT-4.1 classifier over 447 policy documents. The analysis concerns how T\&C rules are represented in the taxonomy, not how they are enforced. 

Two findings emerge. First, high-level category applicability varies sharply by service type: unsafe products account for 17.5\% of e-commerce assignments and 0.2\% of adult-platform assignments, and negative effects on civic discourse reaches 9.9\% within the Meta family but is
absent from adult platforms, Booking.com, and Wikipedia. Second, the categories have notable overlap among sub-category specifications, which makes assigning a single high-level category to a decision difficult. For instance, one of the residual catch-all categories accounts for 21.4\% of assignments and is often assigned alongside a substantive category; the overlap is also visible in the published schema itself. We argue that catch-all reporting is plausibly in part a consequence of the DSA-TDB's schema and attribute design, and recommend a schema that is differentiated by service type with more mandatory attributes derived inductively from
platform governance language.
\end{abstract}

\begin{keyword}
Digital Services Act, Transparency Database, Statement of Reasons, Terms and Conditions,
Content Moderation
\end{keyword}
\end{frontmatter}

% =====================================================================
\section{Introduction}
\label{sec:intro}

Providers of online services make an enormous number of content moderation decisions every day, each marking an exercise of private power over a user's communication. 
Users frequently perceive enforcement as arbitrary and unevenly applied~\cite{duffy2023platform}, and the resulting mistrust extends to platform governance as a
whole~\cite{savolainen2022shadow}.

The Digital Services Act (DSA) responds directly to user mistrust by offering greater transparency into online governance. 
Article 17 requires providers of hosting services to submit a Statement of Reasons (SoR) to any recipient whose content has been restricted, which must be \textit{``clear and easily comprehensible and as precise and specific as reasonably possible under the given circumstances''}~\cite{dsa2022}. 
Article 24(5) requires providers of online platforms to submit every SoR to a publicly accessible database; as a result, the European Commission created the DSA Transparency Database (DSA-TDB), which, since February 2024, has accumulated a continuously updated record of moderation actions across all designated services. 
Each submission conforms to a fixed schema whose most consequential field is a mandatory high-level category attribute with seventeen values that include animal welfare, protection of minors, and illegal or harmful speech. 
Three of the categories are residual (i.e., catch-all groupings) and the remaining fourteen are substantive. 
Because the moderated content itself is excluded for privacy reasons, the category label is the primary signal available to users, regulators and researchers about what kind of rule was enforced when the moderation decision was made.

A substantial body of empirical work reports that the DSA-TDB delivers formal availability of moderation data without substantive transparency. 
Optional attributes intended to support granular reporting go largely unused~\cite{trujillo2025dsa}. 
In addition, free-text explanations are frequently formulaic and machine-generated~\cite{kaushal2024automated}, nearly half of analyzed platforms supply no meaningful information beyond the mandatory minimal categorical label~\cite{groesch2025big}, and the resulting reporting mechanism has not produced interoperable data that can be compared against other reporting mechanisms~\cite{trujillo2026disarranged}. 
The prevailing assessment is one of formal compliance without substantive insight~\cite{klafkowska2025search}.

Current literature has examined the reports and decisions, but not the categories they are classified against. 
Kaushal et~al.~\cite{kaushal2024automated} estimate that over 99.8\% of submitted statements cite a violation of the provider's own terms rather than any legal provision. 
The DSA-TDB can therefore overwhelmingly be considered a record of how platforms apply their own contractual rules, and whether its categories describe those rules well is a question about platforms' terms and conditions. 
Additionally, the same seventeen categories apply to every designated provider irrespective of what the service does, so a social network, a marketplace, a search engine, an encyclopedia and an adult content site all use an identical list to categorize their moderation decisions.

We therefore ask: 
\emph{to what extent do the high-level categories used to classify Statements of Reasons correspond to the content governance rules that providers set out in their terms and conditions, and does that correspondence hold uniformly across service types?} 
We answer this question by mapping the terms and conditions of 31 designated Very Large Online Platforms (VLOPs) and Search
Engines (VLOSEs) onto the DSA-TDB high-level attribute taxonomy, using a few-shot LLM classifier over policy documents. 
Our analysis concerns regulatory representation, i.e., how contractual rules are classified by the taxonomy, and not
moderation behavior.

The paper makes three contributions. 
First, it provides the first systematic mapping between the DSA-TDB taxonomy and the contractual instruments of a large number of designated providers; prior work linking statements to contractual text covers only individual platforms or statements. 
Second, it presents evidence that the taxonomy does not discriminate cleanly over contractual text, including a residual value that behaves as a competing substantive category. 
The core of this finding is visible in the published schema independently
of any classifier or annotator, while the classifier-derived result is weaker and presented with the caveats its validation warrants. 
Third, it argues for differentiating the schema by service type and for the addition of more mandatory attributes to record the kind of rule enforced rather than only an alleged harm.

% =====================================================================
\section{Background and Related Work}
\label{sec:background}

\subsection{The Obligation, the Schema, and the Contractual Basis of Moderation}
\label{subsec:schema}

Art.~3(t) DSA defines \textit{content moderation} as activities taken by online providers to address alleged illegality of content or incompatibility with the provider's own terms and conditions~\cite{dsa2022}. 
Art.~17 DSA requires that an SoR identify which ground applies and explain why it was relied upon, thereby presupposing a traceable correspondence between a moderation action and a clause of the provider's rules.
The governing standard outlined in Art.~17(4) DSA, however, is purely evaluative; \emph{as precise and specific as reasonably
possible} has no benchmark or interpretive guidance, so the required granularity is delegated to the regulated providers.

Submissions conform to a schema of 35 attributes of which two are of notable importance: (i) the decision ground, and (ii) the high-level category~\cite{ec2024dsaapi}. The high-level category is a single value chosen from an exhaustive list of seventeen values, of which fourteen name a type of harmful or illegal content and three are residual. 
Two further attributes, \texttt{category\_addition} and \texttt{category\_specification}, are optional. 
The high-level category values vary in their specificity; some are narrowly delimited, such as \emph{Negative effects on civic discourse} and \emph{Animal welfare}, while others are broad, such as \emph{Violence}. Of the three residual categories, one is a catch-all for any unspecified breach of the provider's terms; however, the use of this term does not require any additional clarification of the contractual rules that have been violated. 

Despite such a lack of clarity, Kaushal et al.~\cite{kaushal2024automated} report that over 99.8\% of SoRs cite this category, meaning that contractual grounds act as the effective default. They also state that the contractual grounds are easier to invoke than statutory frameworks that may vary across Member States, which plausibly explains the asymmetry of contractual grounds used over legal grounds. Similarly, Aspromonte et al.~\cite{aspromonte2024llms} observe that the resulting communication of reasoning for moderation decisions done by online providers is typically generic, meaning that most of what providers submit to the database merely cites a violation of terms and conditions but nothing more. 

\subsection{Reporting Granularity and Its Limitations}

Though a large number of online providers submit daily SoRs to the DSA-TDB, their use varies greatly, which further complicates how moderation decisions are explained. Groesch et al.~\cite{groesch2025big} document inconsistent formats and incomplete reporting; 
Drolsbach and Pr\"ollochs~\cite{drolsbach2024content} report that TikTok performs nearly 350 times more moderation actions per user than X without corresponding differences in the depth of reasoning; and 
Trujillo et al.~\cite{trujillo2025dsa} find the optional granularity attributes seldom used. 
After the Implementing Regulation on transparency reporting~\cite{ec2024templates}, they find platforms still apply divergent procedures and recommend that further attributes be made mandatory~\cite{trujillo2026disarranged}. 
Wei et al.~\cite{wei2024operationalizing} and Sekwenz et al.~\cite{sekwenz2025unfair} attribute such patterns to the DSA's reliance on undefined evaluative concepts. 
Since the moderated content is excluded, and researcher access under Art.~40 DSA has not proved reliable~\cite{leerssen2024outside}, the remaining public resource is the provider's own terms and conditions, which
motivates our approach. 

\subsection{Computational Analysis of Contractual Text}

CLAUDETTE applies machine learning to detect potentially unfair clauses in terms of service against a legally motivated taxonomy~\cite{lippi2019claudette}, an approach since extended to multilingual settings~\cite{galassi2025unfair}.
This work is similar to ours in that it aligns heterogeneous contractual language with a fixed set of legal categories. 
Two studies address the SoR-T\&C relationship directly: 
Aspromonte et al.~\cite{aspromonte2024llms} use large language models to explain SoRs by reference to retrieved sections of the applicable terms; 
E{\ss}er and Spanakis~\cite{esser2025linking} link TikTok's statements to the most relevant clause, annotate 100 statements, and conclude that both are frequently broad, many amounting to catch-all phrasing. 
Both map statements to their explaining text for a single platform; we invert and generalize this, mapping the taxonomy itself onto the terms of all designated providers. 
We also draw on work in legal ontologies, where the granularity of a schema and the relation between its general and specific classes are themselves objects of analysis~\cite{hoekstra2007lkif, sartor2011approaches}; the DSA-TDB taxonomy is a flat list of harms, applied to rules that are not all harm-based.

% =====================================================================
\section{Method}
\label{sec:method}

\subsection{Corpus}
We analyze terms and conditions (T\&Cs) of 31 online services designated as Very Large Online Platforms (VLOPs) or Very Large Online Search Engines (VLOSEs), from the European Commission's DSA Terms and Conditions Database (T\&C Database)~\cite{ec2025tcdb}, a repository of versioned terms and conditions documents of these platforms. 
Where the DSA-TDB records the decisions providers report, the T\&C Database records the documents that define the rules those decisions claim to enforce. 

Retrieval from the T\&C Database yielded 480 documents in July 2026; we filtered for English documents, since we validate the classification for English text, and filtered out purely procedural ones, leaving 447 documents. 
% The corpus is a snapshot since online services update their T\&Cs relatively often. 
We group the 31 services into six types: 
(i) the Meta family (5 services), 
(ii) social media (6),
(iii) e-commerce (6), 
(iv) search and general technology (8), 
(v) adult platforms (4), and 
(vi) other (Booking.com and Wikipedia). 
Document counts are imbalanced, e.g., TikTok Shop has 47 documents and Meta AI only one, so we do not compare aggregate rates between individual platforms.

\subsection{Chunking and Classification}
T\&C documents are chunked hierarchically: the Markdown files are split at headings, so that each chunk belongs to a single policy section and carries a \texttt{section\_path} as metadata. 
The body text is then split at paragraph boundaries and greedily accumulated into chunks up to a limit of 8{,}000 characters. 
The classifier (\texttt{gpt-4.1}) compares each chunk against all 17 categories (3 of which are residual and 14 of which are substantive categories) and returns (i) a match type, (ii) a confidence score, (iii) verbatim evidence quotes, and (iv) a reasoning summary. The prompt supplies the DSA-TDB categories with their official sub-category specifications, and a set of 20 few-shot examples from real T\&C documents, with no training or parameter update. No cross-platform labeled resource that links T\&C provisions to DSA-TDB categories exists (the closest covers 100 statements of a single platform~\cite{esser2025linking}), which is why we apply few-shot learning. 
The match type only distinguishes between \texttt{explicit\_match} and \texttt{no\_match}; each match requires an evidence quote (a section from the T\&C chunk that may have led the model to select a certain match type) and a reasoning summary to help the model ground its classification in the source document. 
Text that merely states that the platform follows a law, describes moderation abstractly, or explains how to report content is to be labeled \texttt{no\_match} unless it names prohibited acts or restricted content. 
The classification yields 6,225 chunk-level high-level category assignments, from a single model and prompt configuration (Section~\ref{sec:limitations}). 
The prompt, few-shot examples, chunking code, model outputs and annotated sample are available from the authors on request.

\subsection{Evaluation Design}
We evaluate against 75 manually annotated chunks, stratified by the classifier's own prediction: 
60 chunks with at least one category assigned, and 15 with none. 
At a corpus-level positive rate below 4\% of chunk\,$\times$\,category cells, random sampling would yield too few positives to estimate precision, while sampling only predicted positives would make recall unmeasurable. 
The annotator, one of the authors without formal legal training, recorded the categories each chunk substantively addresses, blind to the model's labels but aware of the study's framing. 
With a single annotator, inter-annotator agreement cannot be computed; we treat it as a single-reader sanity check, not a gold standard, and no central claim is stated as annotator-validated. 
Scoring is per chunk\,$\times$\,category (1,050 binary predictions over 14 substantive categories); we report precision, recall and F1 rather than accuracy, since \texttt{no\_match} labels would dominate in this heavily imbalanced scenario~\cite{wei2024operationalizing}. 
Since the sample is stratified on the classifier's output, figures are sample-level and not corpus-level estimates, and the 15-chunk negative stratum supports little inference about recall. 
We therefore use the classifier as an exploratory instrument for descriptive purposes and not as a reliable category-level measure.

% =====================================================================
\section{Results}

\subsection{Where the Classifier and the Annotator Disagree}
\label{sec:eval}
First, at the chunk level (whether a passage addresses any category), recall is 1 and precision 0.617. 
In the evaluation, the annotator and classifier agree that 37 chunks address at least one category, and 15 none. 
However, in 23 cases the classifier assigns a category, whereas the annotator assigns none.
The annotator finds in \emph{no} chunks a category that the classifier misses. 
The above rests on only 15 predicted-negative chunks and is weak evidence that the classifier rarely misses text that is relevant to the high-level categories. 

Second, at the assignment level, where we consider each chunk--label pair a separate prediction, precision for \texttt{explicit\_match} is
0.298, recall 0.425 and F1 0.351 (accuracy, at 0.940, is uninformative as only 3.8\% of cells are positive matches). 
Precision below recall means the classifier assigns categories more readily than the annotator, so reported coverage is an upper bound, and since the error directions are independent, absence is interpretable only where the classifier recognizes a category elsewhere in the corpus. 
Precision is highest for categories naming specific conduct, e.g., \emph{Negative effects on civic discourse} 0.75, \emph{Unsafe products} 0.67, and lowest for
the residual \emph{Other violation of provider's terms and conditions} (0.12), which absorbs 23 of the 65 false positives. 
The disagreement admits three readings our design cannot separate: taxonomy indeterminacy, classifier error, and an unreliable single annotation. 
We therefore weigh evidence from the published schema most heavily and treat classifier-derived patterns as suggestive.

\subsection{The Categories Do Not Cleanly Partition Governance Text}
\label{sec:discrim}

\paragraph{Evidence from the published schema.}
The strongest evidence does not depend on any classifier or annotator. 
For instance, the residual category \emph{Other violation of provider's terms and conditions} has sub-category specifications that include \emph{Adult sexual material}, \emph{Nudity}, \emph{Age-specific restrictions} and \emph{Goods/services not permitted to be offered on the platform}. 
These duplicate the subject matter of the high-level categories \emph{Protection of minors} (which specifies \emph{age-specific restrictions concerning minors}), \emph{Unsafe, non-compliant or prohibited products}, and \emph{Cyber violence}. 
On the face of the schema, the residual competes with the substantive categories for material they are defined to describe. 
Moreover, every substantive prohibition in a set of terms is also a violation of those terms, so a provider choosing between a substantive category and the residual chooses between two true descriptions with no rule of choice.

\begin{figure}[tb]
\centering
\includegraphics[width=0.72\linewidth]{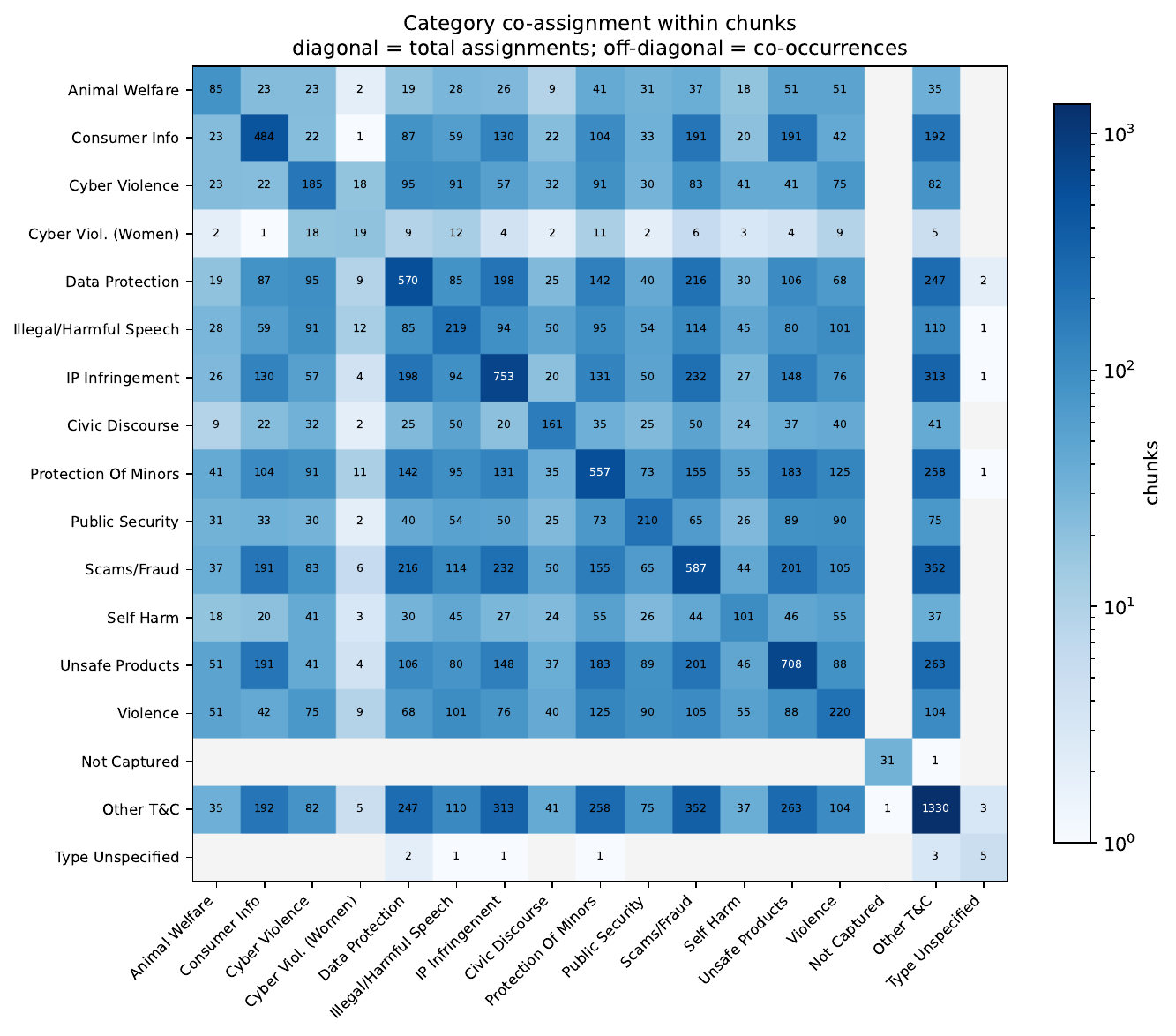}
\caption{Category co-assignment within chunks.
Diagonal cells = total assignments per category; off-diagonal cells = co-occurrences. Gray
indicates no co-assignment.}
\label{fig:coassign}
\end{figure}

\paragraph{Evidence from the classifier's output.}
Figure~\ref{fig:coassign} reports how often categories are assigned to the same chunk. 
A cleanly discriminating taxonomy would produce a strongly diagonal matrix. 
The observed matrix is not: \emph{Scams} co-occurs with \emph{Data protection} 216 times, and with \emph{Unsafe products} 201; 
\emph{Intellectual property} with \emph{Data protection} 198; and
\emph{Protection of minors} with \emph{Unsafe products} 183. 
These counts come from the classifier alone and, given its over-assignment, are likely inflated; they are an upper bound, consistent with but not independent confirmation of the overlap visible in the schema.

The most heavily co-assigned value is a residual category. 
\emph{Other violation of provider's terms and conditions} is assigned 1,330 times, 21.4\% of the total, and is the only value present on all 31 platforms. 
It co-occurs with \emph{Scams} 352 times, \emph{Intellectual property} 313 and \emph{Unsafe products} 263. However, a residual meant for use when no substantive category is relevant should not behave this way.
The other two residual values do behave as residuals; \emph{Not captured by any high-level category} is assigned 31 times and co-occurs once with the catch-all, and \emph{Type of alleged illegal content not specified} is assigned five times. 
Of the uncorroborated assignments to the residual, the annotator mostly assigned a different substantive category; however, these observations also rest on a single annotator since the classifier was not precise for the residual categories. 

\emph{Cyber violence against women} is a limiting case. 
Of its 19 assignments, 18 also carry \emph{Cyber violence}.
The category is a subset of its parent, distinguished only by the gender specification, even though platform terms are drafted in gender-neutral language. 
For example, a Stripchat provision addressing exactly the conduct the category names drew the parent category because nothing in the text supports the narrower one by explicitly addressing women.

\subsection{Coverage}
Table~\ref{tab:coverage} reports total assignments per category. 
No category goes unused, and the median platform is credited with coverage of 12 of the 14 substantive categories; given over-assignment, these
are upper bounds. 
The problem is not that categories are dead but that they are unevenly and, we suggest, ambiguously filled, from 1,330 assignments for the residual to 19 for \emph{Cyber violence against women}.
Coverage correlates moderately with corpus size ($r = 0.43$), but Wikipedia, with nine documents (comparable to Instagram or Pinterest), yields text assigned to only two substantive categories; the
categories absent from its terms (minors, violence, scams) are each assigned on more than 18 platforms
elsewhere, so their absence is unlikely to be a detection failure. 
For an encyclopedia, this suggests that
the categories it must report against largely do not correspond to its own policy instruments.

\begin{table}[tb]
\caption{Chunk-level assignments per category (total 6,225). $^{\dagger}$residual value.}
\label{tab:coverage}
\centering
\small
\begin{tabular}{lr@{\hspace{1.6em}}lr}
\toprule
Category & Assign. & Category & Assign. \\
\midrule
Other violation of T\&Cs$^{\dagger}$ & 1,330 & Violence & 220 \\
Intellectual property & 753 & Illegal or harmful speech & 219 \\
Unsafe or prohibited products & 708 & Public security & 210 \\
Scams and/or fraud & 587 & Cyber violence & 185 \\
Data protection and privacy & 570 & Civic discourse or elections & 161 \\
Protection of minors & 557 & Self-harm & 101 \\
Consumer information & 484 & Animal welfare & 85 \\
Not captured$^{\dagger}$ & 31 & Cyber violence against women & 19 \\
Type not specified$^{\dagger}$ & 5 & & \\
\bottomrule
\end{tabular}
\end{table}

\subsection{Variation by Service Type}

Table~\ref{tab:servicetype} reports the share of each service type's substantive assignments in selected categories, and shows large differences. 
\emph{Unsafe products} accounts for 17.5\% of e-commerce assignments and 0.2\% of adult-platform assignments; \emph{Consumer information} ranges from 22.0\% for the ``Other'' group (Booking.com and Wikipedia) to 1.8\% for adult platforms; and \emph{Negative effects on civic discourse} reaches 9.9\% in the Meta family, but is absent from adult platforms, Booking.com and Wikipedia, which reflects the content rather than a limit on what the classifier can recognize. 
\emph{Data protection} is evenly distributed (16.8--24.5\%), plausibly because it is backed by a mature compliance framework (the GDPR) that standardizes privacy terms across services. 
Groupings are small, so these are descriptive contrasts and not tested differences. 
Additionally, the levels may be inflated by over-assignment, though that does not by itself explain a category being assigned some 87 times more readily for one service type than another. 

\begin{table}[tb]
\caption{Share (\%) of each service type's substantive assignments, with selected categories. Extremes in bold.}
\label{tab:servicetype}
\centering
\small
\begin{tabular}{lrrrrrr}
\toprule
Category & Meta & Social & E-com. & Search & Adult & Other \\
\midrule
Data protection & 13.5 & 9.8 & 10.8 & \textbf{15.0} & 13.9 & \textbf{22.5} \\
Unsafe products & 8.8 & \textbf{15.8} & \textbf{20.5} & 7.2 & \textbf{0.4} & 0.8 \\
Consumer information & 4.0 & 6.3 & 15.2 & 8.1 & \textbf{0.7} & \textbf{28.7} \\
Protection of minors & 13.3 & 13.4 & 7.8 & 9.8 & \textbf{25.6} & 10.9 \\
Civic discourse & \textbf{15.6} & 4.8 & 0.4 & 2.5 & \textbf{0.0} & \textbf{0.0} \\
\midrule
Assignments ($n$) & 377 & 1,625 & 1,806 & 641 & 281 & 129 \\
\bottomrule
\end{tabular}
\end{table}

% =====================================================================
\section{Discussion}

\subsection{A Taxonomy of Harms Applied to Documents That Are Not Only About Harm}

Fourteen of the seventeen values name a type of harmful or illegal content, yet under Art.~3(u) DSA, terms and conditions encompass all clauses governing the contractual relationship, including seller obligations, advertising formats, account eligibility, monetization conditions and developer access. 
In the current taxonomy, there is no value for a decision grounded in a seller-verification requirement or an account-eligibility condition. 
This mismatch is a likely contributor to the pattern in Section~\ref{sec:discrim}: the residual is one of the few values for the body of platform rules that do not concern a harm, and appears to be applied both to provisions its own specifications claim and to commercial or procedural rules for which the taxonomy offers no alternative.

\subsection{Catch-All Reporting and Schema Design}
The finding that 99.8\% of submitted statements are dominated by catch-all labels is ordinarily read as platform behavior: providers taking the least demanding route to compliance. 
However, our results suggest this may not be the full story. 
A classifier with no compliance burden or litigation exposure, given the same seventeen options and documents, also classifies more text with the catch-all category than any substantive category. 
Over-assignment inflates the level and a model-specific bias cannot be ruled out, but neither obviously explains why the inflation lands on this single category, and the overlap in the published specifications offers an independent reason to expect it. 
Catch-all reporting is therefore plausibly in part a consequence of a design that asks providers to classify contractual provisions against a taxonomy of harms, to an undefined standard of specificity, with informative fields made optional~\cite{trujillo2025dsa, kaushal2024automated, groesch2025big, trujillo2026disarranged}. 
However, this does not exclude provider incentives, since our design compares two textual layers and cannot test behavior. 
Nor do we treat the classifier-annotator disagreement as showing that human readers would diverge; it is equally consistent with model or annotation error. 

\subsection{Toward a More Granular Schema}
\label{sec:schema}

We propose four revisions, motivated chiefly by the schema-level evidence and prior literature; they are
hypotheses to test, not conclusions established by our classifier. 
First, a second, orthogonal \emph{mandatory} attribute recording the \emph{kind} of rule enforced, e.g., harm-based, commercial, procedural or eligibility-based, so that harm categories apply only where a harm is at issue, rather than forcing a choice between an ill-fitting harm value and a residual that conveys nothing. 
Its values should be derived inductively from platform governance language and validated with methods from legal ontology engineering~\cite{hoekstra2007lkif, sartor2011approaches}; this study can pose that question but not settle it.

Second, service-type differentiation: service-specific values \emph{added to} a preserved common core, attached to the rule-type attribute, so cross-type comparison is unaffected while within-type comparison becomes possible. 
The objection that this cuts against the harmonization sought by the Implementing Regulation~\cite{ec2024templates} is met in part by Trujillo et al.~\cite{trujillo2026disarranged}, who find platforms still applying divergent
procedures after harmonization, concluding that specifications were clear but compliance was not. 
The trade-off is between a measured cost, up to 21.4\% of assignments in a value that conveys little and providers such as Wikipedia reporting against fourteen categories of which two appear to describe their rules, and a speculative one that a preserved core largely avoids.

Third, informative fields should move from optional to mandatory. For example, \texttt{category\_addition} and \texttt{category\_specification} already offer granularity but providers do not use them~\cite{trujillo2025dsa}; Trujillo et al.~\cite{trujillo2026disarranged} reach the same recommendation from the opposite end of the pipeline. 
Fourth, and at lowest cost, the sub-category specifications require differentiation: faced with the overlap shown in Section~\ref{sec:discrim}, the broader catch-all may be the safer choice for a provider exposed to enforcement risk, so distinguishing the categories in their specifications, not merely their names, is a precondition for the rest.

\section{Limitations}
\label{sec:limitations}

First, the study compares two textual layers and says nothing about enforcement, though a correspondence shows that a provider has articulated a rule addressing a category and not that it is enforced or consistently applied; Wagner et al.~\cite{wagner2024mapping} found written terms accounted for only a fraction of one platform's removals. 
Our conclusions concern the representation of rules in the taxonomy and not moderation behavior.

Second, the reference annotation was produced once, by the first author without legal training, who was aware of the study's framing, so inter-annotator agreement could not be computed and the disagreement in Section~\ref{sec:eval} may reflect annotation or classifier error rather than taxonomy ambiguity. 
Two lines of evidence are not vulnerable to annotation error: 
the duplication among sub-category specifications is a property of the published schema, and the co-assignment structure derives from the classifier alone, albeit subject to its over-assignment; the paper's claims are weighted accordingly. 
Measuring how much of a corpus a category covers would require several legally trained annotators and reported agreement. 

Third, the 75-chunk sample is stratified by the classifier's predictions, so figures are sample-level, recall rests on 15 negative chunks, and precision is low overall (0.298) and for the residual (0.12). 

Fourth, all results derive from one proprietary model and prompt, untested against other models, prompts or annotation strategies. 

Fifth, the analysis is restricted to English-language text and a July 2026 snapshot. 

Finally, the mapping is a structured textual similarity judgment, not a legal determination; it is adequate for exploratory comparison but not for establishing what a provision legally covers.

\section{Conclusion}

We asked whether the categories used to classify Statements of Reasons correspond to the governance rules providers set out in their terms and conditions, and whether that holds across service types. 
Our evidence points toward a negative answer on both parts, with differing levels of strength. 
The overlap between the residual's specifications and substantive categories is observable in the published schema, whereas the classifier-derived patterns are suggestive and limited by modest agreement with a single annotator. 
The problem is not that categories go unused, since every category is assigned somewhere, but that the taxonomy and the documents it is applied to appear to concern partly different subject matter. 
Moreover, the taxonomy's applicability varies by service type even though the reporting obligation does not, and its values overlap enough that assigning a provision to one may often be a choice among plausible options rather than an explicit determination.

The DSA-TDB was intended to make moderation decisions legible to the users they affect. 
However, our results suggest that the known catch-all reporting may follow in part from poor schema design and not solely from unwillingness on the part of providers, though, since we analyze representation and not enforcement, we cannot say how much each factor explains. 
Better reporting is therefore unlikely to come from stricter enforcement of the present schema alone; it may more likely come from a schema that asks providers questions that their own terms can answer.

% =====================================================================
\acknowledgements
David Graus is partly funded by ICAI (AI for Open Government Lab).
Views expressed in this paper are not necessarily shared or endorsed by those funding the research.

\bibliographystyle{vancouver}
\bibliography{references}

\end{document}